\documentclass[letterpaper, 10 pt, conference]{ieeeconf}  

\IEEEoverridecommandlockouts                              

\makeatletter
\let\NAT@parse\undefined
\makeatother

\usepackage[hidelinks]{hyperref}
\usepackage{url}
\let\labelindent\relax
\usepackage[inline]{enumitem}
\usepackage{mathtools}
\usepackage[ruled,linesnumbered,noend]{algorithm2e}
\SetAlgoNoEnd

\usepackage{titletoc} %

\usepackage{setspace}
\usepackage{wrapfig}
\usepackage[most]{tcolorbox}
\usepackage[framemethod=tikz]{mdframed}
\usepackage{listings}

\usepackage{siunitx}
\newcommand{\roundto}[2]{\num[round-mode=places,round-precision=#1]{#2}}

\definecolor{isarblue}{HTML}{006699}
\definecolor{isarfaintblue}{rgb}{0.0, 0.75, 1.0}
\definecolor{isargreen}{HTML}{009966}
\definecolor{red}{HTML}{990000}
\definecolor{patriarch}{rgb}{0.5, 0.0, 0.5}

\lstdefinelanguage{isabelle}{%
    keywords=[1]{type_synonym,datatype,fun,abbreviation,definition,proof,lemma,theorem,qed,corollary,have,hence,also,finally,ultimately,moreover,using,\{},
    keywordstyle=[1]\bfseries\color{isarblue},
    keywords=[2]{where,assumes,shows,fixes,and},
    keywordstyle=[2]\bfseries\color{isargreen},
    keywords=[3]{if,then,else,case,SOME,let,in,O},
    keywordstyle=[3]\color{isarblue},
    keywords=[4]{ATP},
    keywordstyle=[4]\it\color{patriarch},
    keywords=[5]{show,assume,obtain},
    keywordstyle=[5]\bfseries\color{isarfaintblue},
    keywords=[6]{<proof>},
    keywordstyle=[6]\color{yellow},
}

\lstdefinestyle{isabelle}{%
  language=isabelle,
  escapeinside={\&}{&},
  columns=fixed,
  extendedchars,
  basewidth={0.5em,0.45em},
  basicstyle=\singlespacing\ttfamily\small,
  mathescape,
  morecomment=[s][\bfseries\color{red}]{(*}{*)},
  morecomment=[l][\bfseries]{####},
}

\definecolor{mybrown}{RGB}{128,64,0}
\gdef\Sepline{%
  \par\noindent\makebox[\linewidth][l]{%
  \hspace*{-\mdflength{innerleftmargin}}%
   \tikz\draw[thick,dashed,gray!60] (0,0) --%
        (\textwidth+\the\mdflength{innerleftmargin}+\the\mdflength{innerrightmargin},0);
  }\par\nobreak}

\newenvironment{flushitemize}{%
\begin{list}{$\bullet$}
  {\setlength{\leftmargin}{15pt}}%
    \setlength{\labelwidth}{20pt}
    \setlength{\itemindent}{0pt}
    \setlength{\labelsep}{0.5em}
 \setlength{\itemsep}{1pt}
 \setlength{\parskip}{0pt}
 \setlength{\parsep}{0pt}}
 {\end{list}}

\usepackage{amsmath,amsfonts,bm}
\usepackage{stackengine,scalerel,amssymb}

\def\eqref#1{equation~\ref{#1}}

\def\1{\bm{1}}

\DeclareMathAlphabet{\mathsfit}{\encodingdefault}{\sfdefault}{m}{sl}
\SetMathAlphabet{\mathsfit}{bold}{\encodingdefault}{\sfdefault}{bx}{n}

\usepackage[acronym]{glossaries}
\newacronym{gr1}{GR(1)}{Generalized Reactivity(1)}
\newacronym{ltl}{LTL}{Linear Temporal Logic}
\newacronym{sltl}{sLTL}{Simple Linear Temporal Logic}
\newacronym{ec}{EC}{Environment Characterization}
\newacronym{llms}{LLMs}{large language models}
\newacronym{llmrepair}{LLM-Repair}{LLM-Repair}
\newacronym{nl}{NL}{natural languages}
\newacronym{dsl}{DSL}{domain-specific language}
\newacronym{asts}{ASTs}{abstract syntax trees}
\newacronym{er}{ER}{emergency room}

\usepackage[normalem]{ulem}
\usepackage{color}
\usepackage{xcolor}

\newcommand{\call}{\mathcal}

\def\cA{{\call A}}

\def\cD{{\call D}}

\def\cX{{\call X}}
\def\cY{{\call Y}}

\newcommand{\naturals}{\mathbb{N}}
\newcommand{\reals}{\mathbb{R}}

\newcommand{\envspec}{\spec_\textrm{e}}
\newcommand{\sysspec}{\spec_\textrm{s}}

\newcommand{\envinit}{\specinit_\textrm{e}}
\newcommand{\sysinit}{\specinit_\textrm{s}}
\newcommand{\alphainit}{\specinit_\textrm{a}}
\newcommand{\envinitprime}{\specinitprime_\textrm{e}}
\newcommand{\sysinitprime}{\specinitprime_\textrm{s}}
\newcommand{\alphainitprime}{\specinitprime_\textrm{a}}
\newcommand{\envsafety}{\specsafe_\textrm{e}}
\newcommand{\alphasafety}{\specsafe_\textrm{a}}

\newcommand{\envsafetybool}[1]{\specsafe_{\textrm{e}, #1}}
\newcommand{\alphasafetybool}[1]{\specsafe_{\textrm{a}, #1}}
\newcommand{\envsafetyboolprime}[1]{\specsafeprime_{\textrm{e}, #1}}
\newcommand{\alphasafetyboolprime}[1]{\specsafeprime_{\textrm{a}, #1}}
\newcommand{\envsafetyvio}{\varphi_\textrm{violated}}

\newcommand{\syssafety}{\specsafe_\textrm{s}}
\newcommand{\syssafetybool}[1]{\specsafe_{\textrm{s}, #1}}
\newcommand{\syssafetyboolprime}[1]{\specsafeprime_{\textrm{s}, #1}}

\newcommand{\envlive}{\speclive_\textrm{e}}
\newcommand{\alphalive}{\speclive_\textrm{a}}
\newcommand{\envlivebool}[1]{\speclive_{\textrm{e}, #1}}
\newcommand{\envliveboolprime}[1]{\specliveprime_{\textrm{e}, #1}}
\newcommand{\alphalivebool}[1]{\speclive_{\textrm{a}, #1}}
\newcommand{\alphaliveboolprime}[1]{\specliveprime_{\textrm{a}, #1}}
\newcommand{\syslive}{\speclive_\textrm{s}}
\newcommand{\syslivebool}[1]{\speclive_{\textrm{s}, #1}}
\newcommand{\sysliveboolprime}[1]{\specliveprime_{\textrm{s}, #1}}

\newcommand{\spec}{\varphi}
\newcommand{\specinit}{\spec^\textrm{i}}
\newcommand{\specinitprime}{\spec'^{\textrm{i}}}
\newcommand{\specsafe}{\spec^\textrm{t}}
\newcommand{\speclive}{\spec^\textrm{g}}
\newcommand{\specsafeprime}{\spec'^{\textrm{t}}}
\newcommand{\specliveprime}{\spec'^{\textrm{g}}}
\newcommand{\spectask}{\spec_{\textrm{task}}}

\newcommand{\trace}{\tau}

\newcommand{\ap}{\textit{AP}}
\newcommand{\inp}{\cX}
\newcommand{\inpsyms}{{\inp_\textrm{c}}}
\newcommand{\inpuser}{\inp_\textrm{u}}

\newcommand{\ps}{X}

\newcommand{\out}{\cY}
\newcommand{\outnew}{\out_{\textrm{new}}}

\newcommand{\skillnew}{\skill_\textrm{new}}
\newcommand{\skillold}{\skill_\textrm{original}}
\newcommand{\statevar}{\sigma}
\newcommand{\var}{\pi}
\newcommand{\uservar}{\var_\textrm{u}}
\newcommand{\varol}[2]{\var_{{#1}}^{{#2}}}

\newcommand{\skill}{y}

\newcommand{\inpstate}{\statevar_{\inp}}
\newcommand{\inpstatesyms}{\statevar_\textrm{c}}

\newcommand{\outstate}{\statevar_{\out}}

\newcommand{\inpstateprime}{\statevar_{\inp}'}

\newcommand{\vars}{\mathcal{AP}}
\newcommand{\setor}{~|~}

\newcommand{\grounding}{\textrm{G}}

\newcounter{examplecounter}
\newcommand{\examplelabel}[1]{%
  \refstepcounter{examplecounter}%
  \label{#1}%
  \theexamplecounter%
}

\newcounter{problemcounter}
\newcommand{\problemlabel}[1]{%
  \refstepcounter{problemcounter}%
  \label{#1}%
  \theproblemcounter%
}

\newcounter{casecounter}
\newcounter{casecountertwo}
\newcounter{conditioncounter}
\newcounter{defcounter}
\newcommand{\deflabel}[1]{
    \refstepcounter{defcounter}%
    \label{#1}%
    \thedefcounter%
}

\newcommand{\strategy}{\cA_\textrm{s}}
\newcommand{\strategydescription}{\cD}
\newcommand{\counterstrategy}{\cA_\textrm{e}}
\newcommand{\strategystates}{\Sigma}
\newcommand{\strategyinitstates}{\strategystates_0}
\newcommand{\strategytrans}{\delta}
\newcommand{\strategytransenv}{\strategytrans_\textrm{e}}
\newcommand{\strategytranssys}{\strategytrans_\textrm{s}}
\newcommand{\strategylabel}{\gamma}
\newcommand{\strategylabelinp}{\strategylabel_\inp}
\newcommand{\strategylabelout}{\strategylabel_\out}
\newcommand{\strategylabelgoal}{\strategylabel_\textrm{g}}

\newcommand{\framework}{\frameworknospace\space}
\newcommand{\frameworknospace}{\textrm{INPROVF}}

\newcommand{\violation}{t}

\title{\LARGE \bf
INPROVF: 
Leveraging Large Language Models
to 
Repair \\
High-level Robot Controllers 
from Assumption Violations
}

\author{
Qian Meng$^*$, Jin Peng Zhou$^*$, Kilian Q. Weinberger, and Hadas Kress-Gazit%
\thanks{*Equal contribution.}
\thanks{The authors are with Cornell University in Ithaca, NY 14850, USA. Email: {\tt\small \{qm34,jz563,kqw4,hadaskg\}@cornell.edu}. 
}%
}

\begin{document}

\maketitle
\thispagestyle{empty}
\pagestyle{empty}

\begin{abstract}

This paper presents 
\frameworknospace, 
an automatic framework that 
combines
large language models (LLMs) 
and formal methods
to 
speed up the 
repair 
process of 
high-level robot controllers.
Previous approaches
based solely on formal methods
are computationally expensive and cannot scale to large state spaces.
In contrast, 
\framework
uses
LLMs
to 
generate repair candidates, 
and formal methods to verify their correctness.
To improve the quality of these candidates,
our framework 
first
translates
the symbolic 
representations of
the environment
and
controllers 
into natural language descriptions.
If 
a candidate fails the verification, 
\framework provides feedback
on
potential unsafe behaviors 
or unsatisfied tasks,
and
iteratively prompts LLMs to generate improved solutions.
We demonstrate the effectiveness of \framework through 
12 violations with various workspaces, tasks, and state space sizes.
\end{abstract}
\section{Introduction}


Safety-critical robotic applications, 
such as 
autonomous driving~\cite{yurtsever2020survey},
search and rescue missions~\cite{lyu2023unmanned},
and
assistive caregiving robots (e.g.~\cite{madan2024rabbit}),
benefit from formal guarantees to ensure safe operations
and task completions,
the lack of which may result in
harm to users and even
loss of life in catastrophic events.
One way to provide such guarantees is the use of formal synthesis—an automatic technique that transforms temporal logic specifications into correct-by-construction systems— to 
create provably correct high-level robot controllers 
that compose low-level robot skills to 
satisfy temporally extended tasks 
in complex environments~\cite{kress2018synthesis}.


Among formal synthesis methods, 
\gls{gr1} synthesis~\cite{bloem2012synthesis}
is used in robotics applications 
due to its relatively low polynomial-time complexity with respect to the state space 
and its expressivity
covering
many robotic tasks~\cite{menghi2019specification}.
A \gls{gr1} specification 
takes the form of
an implication from \textit{assumptions} about the environment’s behaviors to \textit{guarantees} on the robot's behaviors.
Specifically, the guarantees depend on the validity of
the assumptions;
if any assumption is violated at runtime, 
the synthesized controller is no longer guaranteed to be correct, 
leading to potential undesirable behaviors. 
%


To address assumption violations,
our previous work~\cite{meng2024automated} leverages formal methods-based approaches~\cite{pacheck2022physically} to repair robot behaviors by generating new skills necessary for task satisfaction following assumption violations.
However, our previous repair approach fails to scale to real-world settings with large state spaces at runtime due to the state explosion problem that has challenged formal methods for decades~\cite{valmari1996state}.


Recent advances in \gls{llms} present a unique opportunity to overcome the scalability problem of formal methods-based repair.
Modern \gls{llms}, 
trained on Internet-scale data across diverse domains, 
have strong priors and can address 
complex reasoning tasks--ranging from
competitive mathematics
and
coding problems
to
PhD-level science questions~\cite{openai2025o1}.
Since repairing high-level robot controllers under assumption violation is itself a complex reasoning task, 
we aim to
leverage the reasoning capacities of \gls{llms} to scale up 
repair.

\begin{figure}[t] 
  \centering 
  \includegraphics[width=0.4\textwidth]{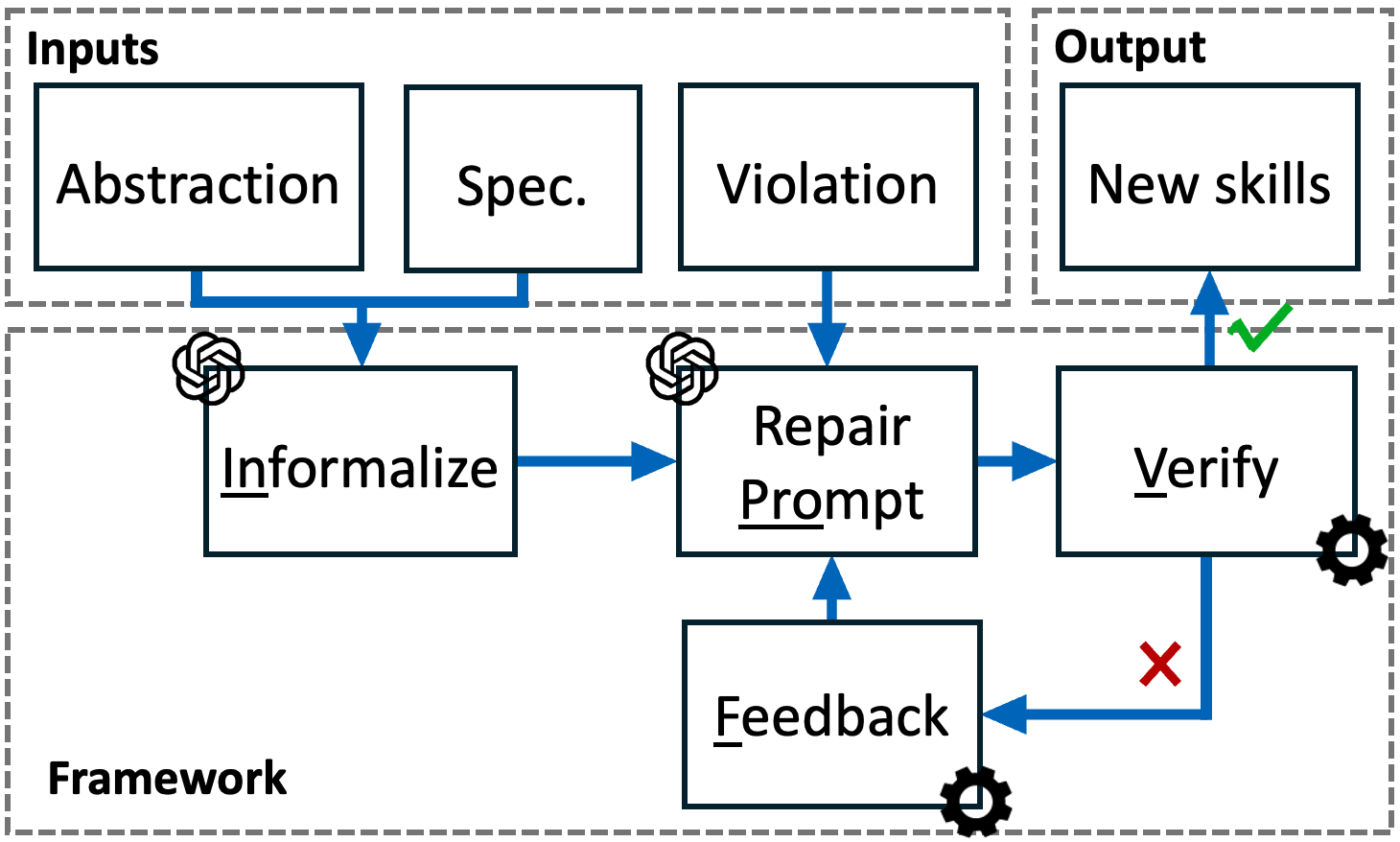} 
  \vspace{-0.1in}
  \caption{\textbf{\framework overview}.
  Given a discrete abstraction, a specification, and an assumption violation 
  we use LLMs and formal methods to repair the behaviors of high-level robot controllers under assumption violations. 
  }
  \label{fig:overview} 
  \vspace{-0.05in}
\end{figure}

In this work, 
we propose \frameworknospace,
a hybrid framework
that 
combines the strengths of \gls{llms} and formal methods
to repair assumption violations in high-level robot controllers at runtime.
We use \gls{llms} to generate repair candidates,
and then verify them through formal methods.
If the verification fails, 
we employ formal analysis to provide feedback,
which is then used by \gls{llms} to 
iteratively refine the candidates.

As illustrated in Fig.~\ref{fig:overview},
\framework consists of
four key steps: 
\underline{In}formalization,
Repair \underline{Pro}mpt, 
\underline{V}erification, 
and \underline{F}eedback.
%
In 
\textit{Informalization},
we convert 
the symbolic abstractions of both the environment 
and the synthesized high-level controller 
into \gls{nl} descriptions that capture their underlying semantics--specifically, the physical meaning of the abstraction and the behavior description of the controller (Sec.~\ref{sec:informalization}).
Our intuition is that \gls{llms},
primarily trained on natural language data and 
further fine-tuned through reinforcement learning from human feedback~\cite{ouyang2022training},
are suitable for processing these informalized \gls{nl} representations.
Next, 
in the \textit{Repair Prompt} step,
we assemble 
the \gls{nl} descriptions of the abstractions, 
the robot's behaviors, 
the task specifications,
and the assumption violations
into a prompt for \gls{llms}
to generate repair candidates--new skills that potentially enable the robot to fulfill its tasks under the violation (Sec.~\ref{sec:repair}).
To ensure the soundness of the candidates,
the \textit{Verification} step first checks their syntax against a predefined grammar,
and then check whether the specification, updated by the violation and the new skills, is realizable (Sec.~\ref{sec:verify}).
If 
the verification step fails,
the \textit{Feedback} step automatically generates \gls{nl} feedback, through syntax and counterstrategy analyses, for \gls{llms} to iteratively generate more promising repair candidates (Sec.~\ref{sec:feedback}).

\textbf{Contributions}.
To our knowledge,
\framework is the first framework that leverages \gls{llms} to accelerate the repair process of assumption violations in large state spaces.
By integrating LLM-based skill generation 
with verification and feedback 
via formal methods, 
\framework enables
provably-correct robotic systems
to remain robust in complex, real-world environments with prohibitively large state spaces.
\section{Related Work}

\textbf{Synthesis for robots}.
Synthesis from \gls{ltl} and \gls{gr1} specifications operates on a discrete abstraction of the environment and the robot's skills~\cite{kress2018synthesis}.
The abstraction can be automatically generated
from sensor inputs using, e.g.~\cite{konidaris2018skills}.
Synthesis takes as input
a temporal logic~\cite{clarke2000model} specification encoding the safety and task requirements for the robot.
While traditionally provided by the user, 
specifications can also be automatically generated from natural language task descriptions using \gls{llms}~\cite{chen2023nl2tl}.
%
Synthesis algorithms (e.g.~\cite{bloem2012synthesis}) automatically generate a high-level controller that composes robot skills to satisfy the given specification, thereby providing correctness guarantees, assuming the abstraction matches the physical world. 

\textbf{Assumption violations}.
%
To address
temporary environment assumption violations, 
\cite{wong2018resilient}
synthesizes robust controllers that can maintain safety constraints during the violations, while
\cite{ehlers2014resilience} tolerates intermittent
violations that occur only a fixed number of times consecutively, separated by violation-free intervals.
Cooperative synthesis
generates controllers that do not attempt to violate assumptions~\cite{ehlers2015synthesizing}.
For long-lasting violations that make the specifications unrealizable, 
\cite{alur2013counter} leverages counterstrategies~\cite{konighofer2009debugging} to generate additional assumptions that restore the realizability.
Similarly, \cite{wong2018resilient} employs
unrealizability analysis~\cite{raman2013towards} to guide the user in adding new liveness assumptions.
Our previous work~\cite{meng2024automated} utilizes synthesis-based repair~\cite{pacheck2022physically}
to create new robot skills to ensure task success given the violations.
%

\textbf{\gls{llms} for formal methods}.
%
The formal methods community has leveraged \gls{llms} to enhance scalability.
\cite{jiang2023draft}
employs \gls{llms} for automated theorem proving, 
tackling mathematical problems that otherwise are computationally intractable and require human inputs.
Similarly,
\cite{pirzada2024llm} 
leverages \gls{llms} to
generate program loop invariants--crucial for reasoning about the correctness of programs with loops--and then verify the invariants using formal techniques.
%
In symbolic execution,
\cite{wang2024python} utilizes \gls{llms} to translate path constraints of real-world Python programs into SMT constraints, improving the scalability of symbolic execution for complex programs.
The software engineering community proposes to
combine \gls{llms} and static analysis to identify software security 
bugs
\cite{li2024llm}.

\section{Preliminaries}
\begin{figure}[t] 
  \centering 
  \includegraphics[width=0.4\textwidth]{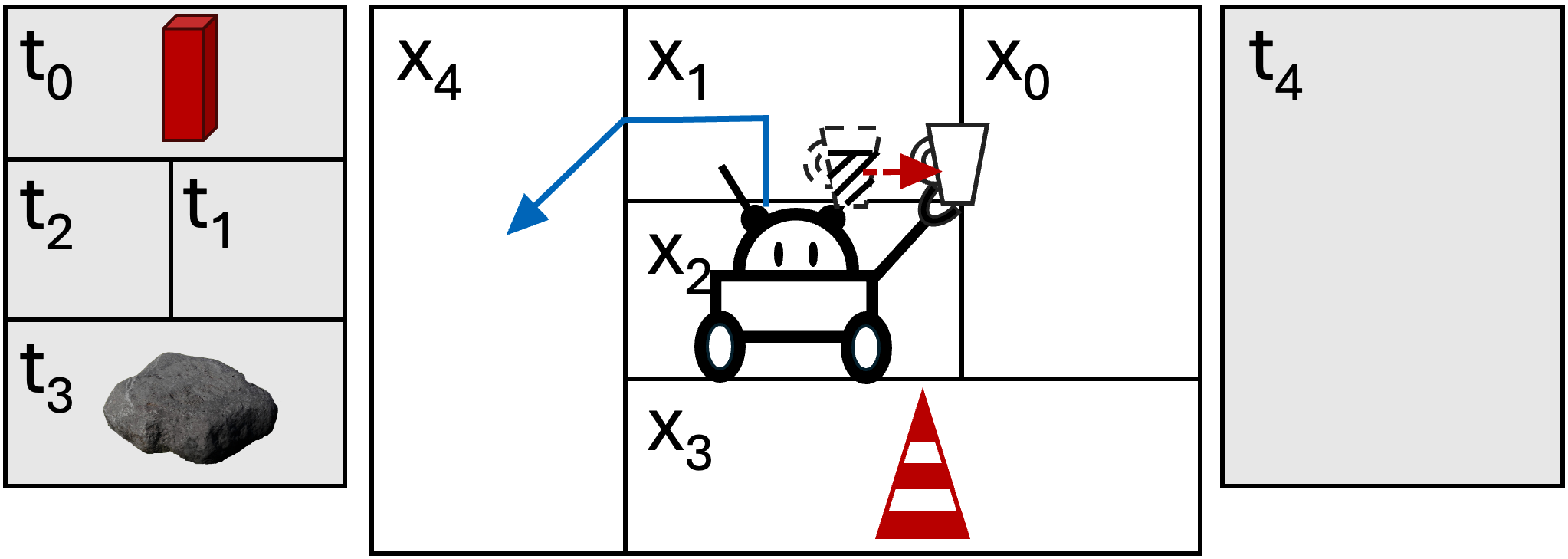} 
  \vspace{-0.1in}
  \caption{\textbf{Example~\ref{example:1} from Sec. VC of~\cite{meng2024automated}}.
  Under an unexpected cup status change,
  the robot creates a new skill 
  (blue arrow) 
  to handle the violation.
  }
  \label{fig:factory_cup} 
\end{figure}
\noindent
\textbf{Example~\examplelabel{example:1}}.
\textit{Consider a mobile manipulator in a factory-like workspace in Fig.~\ref{fig:factory_cup}.
The robot is tasked to move the cup between 
the loading table $t_2$ and 
the assembly table $t_4$ depending on the status of the cup, 
with the assumption that
the status of the cup can only change when it is on the tables,
and the constraint that
an empty cup is not allowed to appear in $x_0$.
However, when the robot holds the cup in $x_2$
while executing a skill $\skillold$ that moves the robot from $x_4$ to $x_0$ via $x_2$,
the cup status changes from full to empty.
The robot should create new skills,
e.g., 
the new skill $\skillnew$ indicated by the blue arrow in Fig.~\ref{fig:factory_cup},
that drives the robot back to $x_4$.
The robot then places the cup to $t_2$,
waits for the cup to be filled up again,
and continues the execution.}

\textbf{Abstractions}.
We model the physical state space with a discrete abstraction. 
An abstraction comprises a set of atomic propositions $\ap$ partitioned into environment inputs $\inp$ and system outputs $\out$ ($\ap = \inp\cup\out$). 
The set of inputs $\inp$, 
representing the environment states,
is further divided into sets of controllable inputs $\inpsyms$ and uncontrollable inputs $\inpuser$ ($\inp = \inpsyms \cup \inpuser$).
Controllable inputs $\inpsyms$
represent the part of the environment that the robot can control indirectly,
e.g. the position of the robot base.
Uncontrollable inputs $\inpuser$
represent the part of the environment that the robot has no control over,
e.g. the position of an obstacle.
%
An input state $\inpstate \subseteq \inp$ and a controllable input state $\inpstatesyms \subseteq \inpsyms$ are subsets of inputs and controllable inputs, respectively.
Let the physical state space be $\ps \subseteq \reals^{n}$.
A grounding function 
$\grounding: \inp \to 2^\ps$ 
maps each input $\var \in \inp$ 
into a set of physical states $\grounding(\var) \subseteq \ps$.
Intuitively, the grounding function captures the physical meaning of each input.

In Example~\ref{example:1},
the inputs are $\varol{o}{r} \in \inp$, 
where $o$ is an object and $r$ is a region,
The grounding function maps each input $\varol{o}{r}$ to the set of physical states 
where the object $o$ locates in the region $r$.
An uncontrollable input $\varol{\textrm{empty}}{}$ represents
that the cup is empty.
The input state in Fig.~\ref{fig:factory_cup} is
$\inpstate \coloneqq \{
\varol{base}{x_2}, 
\varol{cup}{ee}, 
\varol{\textrm{empty}}{}, 
\varol{cone}{x_3}, 
\varol{block}{t_0}, 
\varol{stone}{t_3}
\}$,
meaning
the robot base is in $x_2$, 
the cup is in the end-effector
and is empty,
the cone is in $x_3$,
the block is in $t_0$,
and the stone is in $t_3$.



The system outputs $\out$ represent the robot skills.
Each skill
has a set of preconditions, may 
have intermediate states,
and results in postconditions.
Each state is a controllable input state.
In Example~\ref{example:1},
the precondition of the new skill $\skillnew$ is
$\{\varol{base}{x_2}, 
\varol{cup}{ee}, 
\varol{block}{t_0}
\}$ (stone and cone are uncontrollable),
the intermediate state is
$\{\varol{base}{x_1}, 
\varol{cup}{ee}, 
\varol{block}{t_0}\}$,
and the postcondition is
$\{\varol{base}{x_4}, 
\varol{cup}{ee}, 
\varol{block}{t_0}
\}$.

\textbf{Specifications}.
We use the \gls{gr1} fragment of \gls{ltl}
for its synthesis algorithm and expressivity.
%
The syntax of \gls{ltl} is
$\spec \coloneqq  \pi \setor \neg\spec \setor \spec \vee \spec \setor$$\bigcirc \spec \setor$$\spec\ \mathcal{U} \spec$
where
a formula $\spec$ is constructed recursively from atomic propositions $\pi\in\vars$,
logical operators ($\neg, \vee$),
and temporal operators \textit{next} ($\bigcirc$) and \textit{until} ($\mathcal{U}$).
Additional logical operators ($\land, \to, \leftrightarrow$)
and temporal operators \textit{eventually} ($\lozenge$) and \textit{always} ($\square$) 
can be derived from previous operators.
The semantics of \gls{ltl} is defined over infinite traces $\trace = (\statevar_0, \statevar_1, \dots)$, where each $\statevar_i \subseteq \vars$ denotes the set of propositions that are true at step $i$. When a trace suffix starting at position $i$ satisfies a formula $\spec$, we denote this as $\trace^i \models \spec$. 
By convention, we say that a trace $\trace$ satisfies a formula $\spec$, denoted as $\trace \models \spec$, if the formula is satisfied from the beginning of the trace (i.e., $\trace^0 \models \spec$). 
The satisfaction relation is defined recursively as follows: $\trace^i \models \pi$ if $\pi \in \statevar_i$ ($\pi$ is true at step $i$); 
$\trace^i \models \bigcirc\spec$ if $\trace^{i+1} \models \spec$ ($\spec$ holds at the next step); $\trace^i \models \lozenge\spec$ if $\exists j \ge i$ such that $\trace^j \models \spec$ ($\spec$ eventually holds); and $\trace^i \models \square\spec$ if $\forall j \ge i$, $\trace^j \models \spec$ ($\spec$ always holds).
A more detailed introduction of \gls{ltl} semantics can be found in~\cite{clarke2000model}.

Specifications in the \gls{gr1} fragment of \gls{ltl} take the form 
$\spec = \envspec \to \sysspec$
where 
$\envspec = \envinit \land \envsafety \land \envlive$ represents the environment assumptions and 
$\sysspec = \sysinit \land \syssafety \land \syslive$ denotes the system guarantees.
The initial constraints $\envinit$ and $\sysinit$, 
defined over $\inp$ and $\out$, 
specify the initial conditions of the environment and the system, respectively.
The safety constraints,
$\envsafety$ and $\syssafety$,
capture conditions that must hold at all times
and take the form of 
$\alphasafety = \bigwedge_i \square \alphasafetybool{i}$,
for $\textrm{a} \in \{\textrm{e, s}\}$,
where $\envsafetybool{i}$ is defined over $\vars\cup \bigcirc \inp$
and $\syssafetybool{i}$ over $\vars\cup \bigcirc \vars$.
%
%
Lastly, we have the liveness constraints $\envlive$ and $\syslive$
which encode
environment assumptions and system goals that must be satisfied repeatedly.
For $\textrm{a} \in \{\textrm{e, s}\}$,
$\alphalive = \bigwedge_i\square\lozenge \alphalivebool{i}$
where each $\alphalivebool{i}$ is defined over $\vars$.

Writing \gls{gr1} specification can be partially automated from the abstraction (e.g., from topological constraints, skills definitions, etc), allowing the user to provide the  
task specification $\spectask$ that captures the essence of the task. 
We define such $\spectask$ as a \textit{sub-specification} of $\spec$.

%

\noindent\textbf{Definition~\deflabel{def:subspec}.} 
Let $\spec = \envinit \land \bigwedge_i \square \envsafetybool{i} \land \bigwedge_i\square\lozenge \envlivebool{i} \to \sysinit \land \bigwedge_i \square \syssafetybool{i} \land \bigwedge_i\square\lozenge \syslivebool{i}$ 
and 
$\spectask = \envinitprime \land \bigwedge_i \square \envsafetyboolprime{i} \land \bigwedge_i\square\lozenge \envliveboolprime{i} \to \sysinitprime \land \bigwedge_i \square \syssafetyboolprime{i} \land \bigwedge_i\square\lozenge \sysliveboolprime{i}$ 
be \gls{gr1} specifications.
$\spectask$ is a \textit{sub-specification} of $\spec$, denoted $\spectask \preceq \spec$, 
if $\forall \textrm{a} \in \{\textrm{e, s}\}$,
$\alphainitprime = \alphainit$,
$\{\alphasafetyboolprime{i}\}_i \subseteq \{\alphasafetybool{i}\}_i$,
$\{\alphaliveboolprime{i}\}_i \subseteq \{\alphalivebool{i}\}_i$.

In Example~\ref{example:1},
an example constraint not included in
the task specification
$\spectask$ but in the full \gls{gr1} specification $\spec$ is:
one object cannot appear in two places simultaneously.




\textbf{Synthesis}.
\gls{gr1} synthesis~\cite{bloem2012synthesis} first transforms the \gls{gr1} specification into a two-player game between the environment and the system, 
where the transition rules for each player are defined by the safety constraints $\envsafety$ and $\syssafety$,
and the winning conditions are defined by the liveness goals $\envlive$ and $\syslive$, respectively.
Next, synthesis computes a set of winning states from which the system is guaranteed to win the game, a process called \textit{realizability check}. 
Finally, synthesis extracts a finite automaton from the winning states as the robot \textit{strategy}.

\noindent \textbf{Definition~\deflabel{def:strategy}.}
A \textit{strategy} is a
deterministic finite automaton $\strategy = (\strategystates, \strategyinitstates, \inp, \out, \strategytrans, \strategylabelinp, \strategylabelout)$ where 
\begin{flushitemize}
    \item $\strategystates$ is a set of states and $\strategyinitstates \subseteq \strategystates$ is the set of initial states. 
    \item $\inp$ and $\out$ are sets of inputs and outputs.
    \item $\strategytrans: \strategystates \times 2^\inp \to \strategystates$ 
    is the transition function that 
    maps a state and an input state to the next state. 
    \item $\strategylabelinp: \strategystates \to 2^\inp$
    maps a state to its labeled input state.
    \item $\strategylabelout: \strategystates \to 2^\out$ 
    maps a state to its labeled output state. 
\end{flushitemize}

A \textit{strategy description} $\strategydescription$ for the strategy $\strategy$ is a natural language description of how $\strategy$ would behave.

If the set of winning states does not overlap with all system liveness goals or does not cover the initial states,
a winning strategy for the system does not exist,
i.e. the specification is unrealizable.
Conversely, a winning strategy for the environment, or a \textit{counterstrategy}~\cite{konighofer2009debugging}, exists.

\noindent \textbf{Definition~\deflabel{def:counterstrategy}.}
A \textit{counterstrategy} is a nondeterministic finite automaton $\counterstrategy = (\strategystates, \strategyinitstates, \inp, \out, \strategytransenv, \strategytranssys, \strategylabelinp, \strategylabelout, \strategylabelgoal)$ where
\begin{flushitemize}
    \item $\strategystates, \strategyinitstates, \inp, \out, \strategylabelinp, \strategylabelout$ are the same as in Definition~\ref{def:strategy}.
    \item $\strategytransenv: \strategystates \to 2^\inp$ 
    maps a state 
    to the next input state.
    \item $\strategytranssys: \strategystates \times 2^\inp \to 2^\strategystates$ 
    maps a state 
    $\statevar$ 
    and the next input state 
    $\inpstate$ 
    to a set of possible next states
    $\strategytranssys(\statevar, \inpstate) \subseteq 2^\inp$.
    \item $\strategylabelgoal: \strategystates \to \naturals$ 
    maps a state $\statevar$ to the index of a system liveness goal the state attempts to satisfy.
\end{flushitemize}
%
When synthesis produces a strategy, 
we deploy the strategy on robots as the high-level controller.
During the robot execution, 
environment safety assumptions $\envsafety$ may be \textit{violated}.

\noindent
\textbf{Definition~\deflabel{def:violation}.}
An \textit{assumption violation} 
is a triplet $\violation = (\inpstate, \outstate, \inpstateprime)$ such that
$\exists i$, $\inpstate \cup \outstate \cup \bigcirc \inpstateprime \not\models \envsafetybool{i}$.

In Example~\ref{example:1},
$\inpstate \coloneqq \{
\varol{base}{x_2}, 
\varol{cup}{ee},  
\varol{cone}{x_3}, 
\varol{block}{t_0}, 
\varol{stone}{t_3}\}$,
$\outstate \coloneqq \{\skillold\}$,
and 
$\inpstateprime \coloneqq 
\inpstate \cup \{\varol{\textrm{empty}}{}\}$.
\section{Problem Statement}
We tackle the runtime repair problem of high-level robot controllers under assumption violations.

\noindent
\textbf{Problem~\problemlabel{problem:runtime_repair}.}
Given 
\begin{enumerate*}[label=(\roman*),ref=\roman*]
    \item an abstraction of the environment $\inp$ and skills $\out$,
    
    \item specifications consisting of a task specification $\spectask$ and a GR(1) specification $\spec$ such that $\spectask \preceq \spec$,
    
    \item an assumption violation $\violation = (\inpstate, \outstate, \inpstateprime)$,
    and
    \item a set of example strategies and their descriptions 
    $\{(\strategy^i, \strategydescription^i)\}_{i=1}^k$;  
\end{enumerate*}
find a set of new skills $\outnew$ that makes the specification $\spec$ realizable under the violation $\violation$, at runtime.
\section{Approach}
\begin{figure}[t] 
  \centering 
  \includegraphics[width=0.486\textwidth]{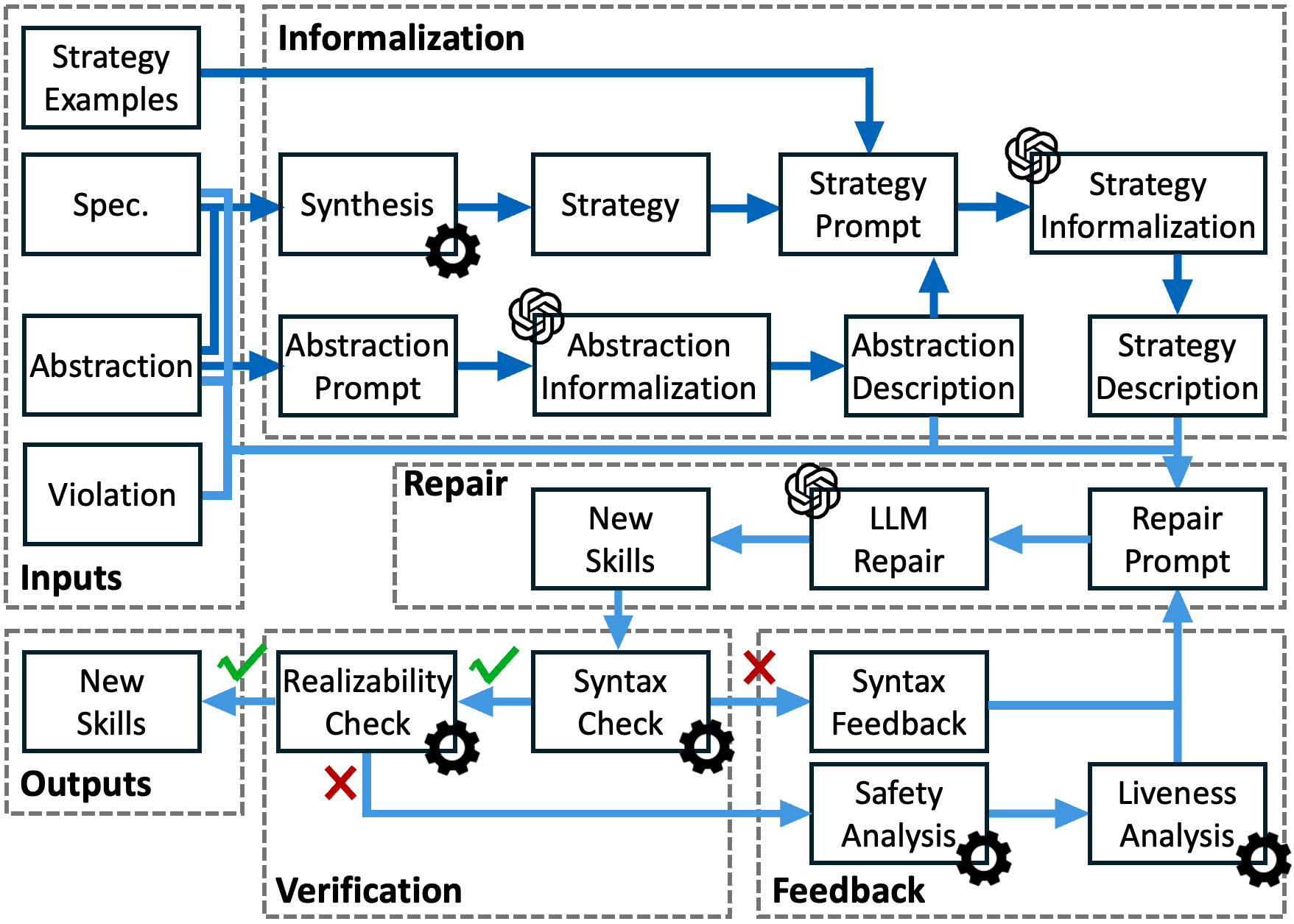} 
  \vspace{-0.25in}
  \caption{\textbf{\framework system details}.
  Our system 
  consists of four steps: 
  \textit{Informalization} translates formal abstraction and strategy into natural language descriptions; 
  \textit{Repair} leverages \gls{llms} to generate new skill candidates; 
  \textit{Verification} ensures the correctness of the candidates through syntax and realizability checking;
  \textit{Feedback} leverages formal analyses to provide concise \gls{nl} feedback to assist \gls{llms} for iterative prompting.
  }
  
  \label{fig:framework} 
  \vspace{-0.2in}
\end{figure}

To address
Problem~\ref{problem:runtime_repair} in large state spaces, 
we propose a hybrid framework \frameworknospace, 
as depicted in Fig.~\ref{fig:framework}, 
that
leverages \gls{llms} to efficiently produce new
skill candidates, and formal methods to verify them.
We first informalize the abstraction into 
the \gls{nl} description of its physical meaning.
We then synthesize a strategy from the specification and
use few-shot prompting to obtain the \gls{nl} description of the strategy behavior
(Sec.~\ref{sec:informalization}). 
Next, we leverage \gls{llms} to generate a set of new skill candidates that potentially repair the behavior of the strategy under the violation
(Sec.~\ref{sec:repair}).
We verify the candidates through syntax and realizability checking
(Sec.~\ref{sec:verify}). 
If the candidates cannot repair the violation,
we provide feedback on
syntax and counterstrategy
analyses to the \gls{llms}
(Sec.~\ref{sec:feedback}). 
Fig.~\ref{fig:casestudy} 
presents
a case study 
that walks through our framework with Example~\ref{example:1}.

\begin{figure*}[t]
  \centering
  \begin{minipage}[t]{0.496\textwidth}
  \vspace{0pt}
    \begin{tcolorbox}[
      colback=blue!5!white,      
      colframe=blue!75,
      width=\textwidth,
      boxrule=0.5pt,
      sharp corners
    ]
    \scriptsize\textbf{A. Abstraction informalization prompt.} \\
You are a researcher in robotics and an expert in formal methods for robotics.
You are given an abstraction that contains a set of input propositions X, a physical state space PS \verb|\|subseteq R\verb|^|n, and a grounding function G: X \verb|->| 2\verb|^|PS that maps each input proposition to a set of physical states. You are asked to explain the physical meaning of the input propositions.

\# Input propositions: 
\textbf{\textless Inserted input propositions $\inp$ \textgreater}

\# Grounding: 
\textbf{\textless Inserted grounding function $\grounding$ \textgreater}

Please only provide a concise, general explanation covering all input propositions' physical meaning. Do not provide any examples.

\textbf{\textless Informalized abstraction to be returned here\textgreater}
    \end{tcolorbox}
    \vspace{-1.5em}
    \begin{tcolorbox}[
      colback=red!5,       
      colframe=red!75,
      width=\textwidth,
      boxrule=0.5pt,
      sharp corners
    ]
    \scriptsize\textbf{B. Example informalized abstraction description from \gls{llms}.} \\
    Each input proposition \verb|\( p_{o\_r} \)| represents the condition that object \verb|\( o \)| is located within region \verb|\( r \)|. The grounding function \verb|\( G \)| maps each proposition \verb|\( p_{o\_r} \)| to the set of physical states where the pose of object \verb|\( o \)| is in region \verb|\( r \)|. The proposition "empty" signifies the status of the cup, indicating whether it is empty.
    \end{tcolorbox}
\vspace{-1.5em}
    \begin{tcolorbox}[
      colback=blue!5!white,      
      colframe=blue!75,
      width=\textwidth,
      boxrule=0.5pt,
      sharp corners
    ]
    \scriptsize\textbf{C. Strategy informalization prompt.} \\
    You are a researcher in robotics and an expert in formal methods for robotics.
You are given an abstraction of the environment, a set of robot skills, a GR(1) specification, and its corresponding synthesized strategy. You are asked to describe the behavior of the strategy. I will first give an example of a strategy, its behavior description, and its explanation. Then, I will ask you to describe the behavior of a different strategy.

\# Example: \textbf{\textless Inserted strategy informalization example
$(\strategy^1, \strategydescription^1)$. 
See Fig.~\ref{fig:inform_example}A for the exact text \textgreater}

\# Problem:
Now, I will ask you to describe the behavior of a different strategy. You are given: (1) an abstraction of the physical workspace and objects in the workspace, (2) a set of robot skills, (3) a task specification, and (4) a synthesized robot controller.

\#\# Input propositions:
\textbf{\textless Inserted inputs $\inp$ and informalized abstraction\textgreater}

\#\# Output propositions:
\textbf{\textless Inserted output propositions $\out$\textgreater}

\#\# Task specification:
We use the \gls{gr1} fragment of Linear Temporal Logic to encode the task specification. The task specification is given in the following JSON file:
\textbf{\textless Inserted task specification $\spectask$\textgreater}

\#\# Strategy:
We represent the robot strategy synthesized from the specification as a finite state automaton. 
For each node, 'rank' represents which liveness goal the node is trying to satisfy, 'state' represents the value of each variable in the state, in the order of the field 'variables', and 'trans' represents the possible successor states. 
The strategy is in the JSON file:
\textbf{\textless Inserted strategy $\strategy$\textgreater}

Please describe the behavior of the synthesized strategy only. Do not provide any explanation:
\textbf{\textless Informalized strategy behavior to be returned here\textgreater}
    \end{tcolorbox}
\vspace{-1.5em}
        \begin{tcolorbox}[
      colback=red!5,       
      colframe=red!75,
      width=\textwidth,
      boxrule=0.5pt,
      sharp corners
    ]
    \scriptsize\textbf{D. Example informalized strategy behavior from \gls{llms}.} \\
The synthesized strategy behaves as follows:

Initially, the robot's base is at position x4, the cup is at location t2, the block is at location t1, the cone is at x3, the stone is at t0, and the cup is empty.

When the cup becomes **not empty**, the robot executes **skill4** to pick up the block from t1 and then **skill7** to place the block at t3. Next, the robot uses **skill2** to pick up the cup from t2. It then executes **skill1** to move from x4 via x2 to x0 while holding the cup. At x0, the robot uses **skill6** to place the cup on the table.

When the cup becomes **empty** again, the robot executes **skill3** to pick up the cup from the table. It then uses **skill0** to move from x0 via x2 to x4 while holding the cup. At x4, the robot executes **skill5** to place the cup back at t2.

This behavior repeats indefinitely, with the robot moving the cup between t2 and the table whenever the cup transitions between empty and not empty.

    \end{tcolorbox}
\vspace{-1.5em}
    \begin{tcolorbox}[
      colback=red!5,       
      colframe=red!75,
      width=\textwidth,
      boxrule=0.5pt,
      sharp corners
    ]
\scriptsize\textbf{G. Example new skill candidates that cannot satisfy liveness constraints.}
\begin{lstlisting}[basicstyle=\scriptsize]
```JSON
{   "new_skill_0": [
        [["p_base_x2", "p_cup_ee", "p_block_t3"], 
        [["p_base_x4", "p_cup_ee", "p_block_t3"]]]]}
```
\end{lstlisting}
\textbf{Feedback from liveness analysis}: 
The new skills cannot satisfy the liveness goals 
( empty \verb|->| p\_cup\_t2 )
    \end{tcolorbox}
    
  \end{minipage}
  \hfill
  \begin{minipage}[t]{0.496\textwidth}\vspace{0pt}
    
        \begin{tcolorbox}[
      colback=blue!5!white,      
      colframe=blue!75,
      width=\textwidth,
      boxrule=0.5pt,
      sharp corners
    ]
    \scriptsize\textbf{E. Repair prompt.} \\
    You are a researcher in robotics and an expert in formal methods for robotics.
You are given: (1) an abstraction of the physical workspace and objects in the workspace, (2) a set of robot skills, (3) a task specification, (4) the original behavior of the synthesized strategy, and (5) a violated observation during the robot execution.
You are asked to provide a set of new skills that along with the original skills can allow the robot to recover from the violation and satisfy the task specification.

\# Abstraction:\\
\#\# Input propositions:
\textbf{\textless Inserted inputs $\inp$ and informalized abstraction\textgreater}

\#\# Output propositions:
\textbf{\textless Inserted output propositions $\out$\textgreater}

\# Task specification:\\
We use the \gls{gr1} fragment of Linear Temporal Logic to encode the task specification. The task specification is given in the following JSON file:
\textbf{\textless Inserted task specification $\spectask$\textgreater}

\# Strategy behavior:
\textbf{\textless Inserted informalized strategy behavior\textgreater}

\# Assumption Violation:
During the robot execution, we detect that an observed transition violates some environment safety assumptions.

\#\# The violated observation is in the following JSON file:\\
\textbf{\textless Inserted assumption violation $t=(\inpstate, \outstate, \inpstateprime)$\textgreater}

\#\# The violated environment safety assumptions are as follows:\\
\textbf{\textless Inserted violated assumptions $\envsafetyvio$ \textgreater}

\# Returns:
After the assumption violation, we relax the violated assumption to include the observed violation transition. You are asked to provide a set of new skills that, along with the original skills, allow the robot to continue satisfying the task after the assumption violation.
Please respond with new skills only in the following format:
\begin{lstlisting}[basicstyle=\scriptsize]
```JSON
{  "new_skill_0": [
            [ <PRECONDITION>, 
                [<POSTCONDITION>, ...]], 
            [ <PRECONDITION>, 
                [<POSTCONDITION>, ...]], 
            ...],
    "new_skill_1": [
            [ <PRECONDITION>, 
                [<POSTCONDITION>, ...] ], 
            [ <PRECONDITION>, 
                [<POSTCONDITION>, ...]], 
            ...],
    ...}
```
\end{lstlisting}
The response follows the grammar:
\begin{scriptsize}
\begin{verbatim}
<new_skill>: [<intermediate_transition>+]
<intermediate_transition> = [<precondition>, 
                            [<postcondition>+]]
<precondition> = [<controllable_input>+]
<postcondition> = [<controllable_input>+]
\end{verbatim}
\end{scriptsize}

Each pre or post-condition in the new skills should be a complete controllable input state.
Do not include original skills in the new skills. 
Do not include any comments in skills.
\textbf{\textless New skill candidates to be returned here\textgreater}
    \end{tcolorbox}
    \vspace{-1.5em}
    \begin{tcolorbox}[
      colback=red!5,       
      colframe=red!75,
      width=\textwidth,
      boxrule=0.5pt,
      sharp corners
    ]
\scriptsize\textbf{F. Example new skill candidates that violate safety constraints.}
\begin{lstlisting}[basicstyle=\scriptsize]
```JSON
{    "new_skill_0": [
        [["p_base_x2", "p_cup_ee"],
        [["p_base_x3", "p_cup_ee"]]],
        [["p_base_x3", "p_cup_ee"],
        [["p_base_x4", "p_cup_ee"]]]]}
```
\end{lstlisting}
\textbf{Feedback from safety analysis}: 
new\_skill\_0 violates the hard constraints 
!(p\_base\_x3' \& p\_cone\_x3')
    \end{tcolorbox}

  \end{minipage}
  
  \caption{\textbf{Case study for Example~\ref{example:1}}. 
  (A)-(D) 
  describe
  the prompts and 
  LLM responses from informalization.
  (E)-(G)
  present the
  repair prompt and 
  two sets of new skill candidates from \gls{llms},
  along with the feedback from formal analyses.
  We use \textrm{p\_object\_region}
  to represent the input $\varol{\textrm{object}}{\textrm{region}}$ and prime (') for the next operator ($\bigcirc$), following the syntax of \gls{gr1} formulas in Slugs~\cite{ehlers2016slugs}.
%
  }
  \label{fig:casestudy}
  \vspace{-0.25in}
\end{figure*}

\subsection{Informalization}\label{sec:informalization}
We begin by leveraging \gls{llms} to translate 
the environment abstraction and the synthesized strategy into informal natural language descriptions. 
Since
\gls{llms} are primarily trained on natural language data and optimized for a wide range of reasoning tasks on \gls{nl} traces,
we hypothesize that \gls{llms} better utilize the abstraction and strategy behaviors in natural languages than in the formal representations.

\textbf{Abstraction informalization}.
We take
in an abstraction consisting of 
the environment inputs $\inp$ and 
a grounding function $\grounding$ 
that maps each input to a set of physical states.
We prompt \gls{llms} to describe the physical meaning of the abstraction. 
Fig.~\ref{fig:casestudy}A
presents the prompt,
and Fig.~\ref{fig:casestudy}B
shows the informalized abstraction description 
for Example~\ref{example:1}.
%
%

\textbf{Strategy informalization}.
Next, we 
obtain
the behavior description of the strategy $\strategy$, 
synthesized from the \gls{gr1} specification $\spec$.
Since each state of the strategy consists of environment input propositions, 
we leverage the previously generated environment abstraction description to 
provide the context for \gls{llms} to ingest the strategy behavior.
Our preliminary study found that this is a relatively challenging task
due to the complexity of the strategy, which is a finite automaton as described in Definition~\ref{def:strategy}.
So,
we use one-shot prompting--including one example translation in the prompt--to help steer \gls{llms} towards generating accurate strategy behavior.
Fig.~\ref{fig:casestudy}C
describes the one-shot prompt
and
Fig.~\ref{fig:casestudy}D
shows the resulting strategy behavior 
for Example~\ref{example:1}.
Fig.~\ref{fig:inform_example}A shows the example translation for Example~\ref{example:1}.

\subsection{Repair}\label{sec:repair}

The repair process takes in 
the description of the abstraction $\inp$,
the existing robot skills $\out$,
the task specification $\spectask \preceq \spec$, 
provided by the user,
that only contains the essential constraints of the full \gls{gr1} specification $\spec$, 
the assumption violation $t = (\inpstate, \outstate, \inpstateprime)$,
and the behavior description $\strategydescription$ of the strategy $\strategy$ 
synthesized from $\spec$.
We form a prompt that combines the inputs and asks \gls{llms}
to only produce a set of new skill candidates $\outnew$ 
in a predefined 
\gls{dsl}.
%
We also require the new skills to be formatted as JSON files only,
and we perform regular expression matching to extract $\outnew$ 
from the responses.
Fig.~\ref{fig:casestudy}E
presents the repair prompt,
and Figs.~\ref{fig:casestudy}F and~\ref{fig:casestudy}G 
show two example repair candidates
for Example~\ref{example:1}.


\subsection{Verification}\label{sec:verify}

After obtaining a set of new skill candidates $\outnew$,
we verify the correctness of $\outnew$ 
through syntax and realizability checking, 
ensuring that $\outnew$ 
indeed
repairs the violation.

\textbf{Syntax check}.
We check the syntax of $\outnew$ against 
the \gls{dsl} grammar 
shown in the repair prompt 
(Fig.~\ref{fig:casestudy}E).
We leverage a parser to automatically transform $\outnew$ into a set of \gls{asts}.
If parsing is successful, 
the new skills are in the correct syntax.
We also ensure type correctness by 
recursively traversing each node of the \gls{asts}, 
verifying each component of $\outnew$ against its expected type.

\textbf{Realizability check}.
Once $\outnew$ is in the correct syntax, 
we check whether $\outnew$ can repair the assumption violation.
We use a compiler from~\cite{meng2024automated} to add $\outnew$ to $\spec$ and relax the assumptions in $\spec$ that are violated by the violation 
$t = (\inpstate, \outstate, \inpstateprime)$, 
resulting in an updated specification $\spec'$.
We then leverage the realizability check of a \gls{gr1} synthesizer Slugs~\cite{ehlers2016slugs} to check 
the realizability of $\spec'$.
If $\spec'$ is realizable,
then
$\outnew$ is verified to be sufficient to solve the violation.
\subsection{Feedback}\label{sec:feedback}
If the new skills $\outnew$ do not pass the verification step,
we provide three types of feedback,
each of which targets a different type of error,
to ensure that we have informative
and yet concise \gls{nl} feedback to \gls{llms} for iterative prompting.

\textbf{Syntax feedback}. For syntax errors detected in the syntax-checking procedure, 
we provide feedback based on the nodes of the ASTs that violate the grammar and the grammar rules that are violated.
For instance, 
for a new skill $\skillnew \in \outnew$ that contains an uncontrollable input $\uservar\in\inpuser$ in its precondition,
we automatically provide the following feedback: ``$\uservar$ in the precondition of $\skillnew$ is not a controllable input."

\textbf{Counterstrategy}. Next, if the new skills $\outnew$ are syntactically correct but the updated specification $\spec'$ remains unrealizable, 
meaning skills in $\outnew$ do not repair the violation,
we compute a counterstrategy $\counterstrategy$
to provide feedback.
While the counterstrategy contains information about the cause of the unrealizability of $\spec'$,
\gls{llms} may not be able to utilize the formal representation of the counterstrategy (as shown in Definition~\ref{def:counterstrategy}).
We thus perform formal analysis on the counterstrategy to 
transform it 
into concise \gls{nl} feedback.


\textbf{Safety analysis}.
Given the counterstrategy $\counterstrategy$ and
the system safety constraints $\syssafety = \bigwedge_i \square \syssafetybool{i}$,
we provide feedback on if and which skill violates any safety constraint $\square\syssafetybool{i}$.
As shown in Algorithm~\ref{alg:safety_analysis},
We first identify any transition $(\statevar, \statevar')$ in $\counterstrategy$ such that $\statevar'$ is a sink state, i.e. there are no outgoing edges from $\statevar'$ (Line~\ref{line:identify sink state}).
We are interested in sink states because those are the states where the system has no valid next move, indicating states that violate safety constraints. 
We then iterate through all system safety constraints $\square \syssafetybool{i}$ (Line~\ref{line:safety}), 
and check whether the transition violates any safety constraint
(Line~\ref{line:sat_check}).
We perform this check by 
formulating a conjunction of 
the labels of $\statevar$, 
the labels of $\statevar'$,
and the safety constraint of interest,
and then leveraging an SMT solver 
Z3~\cite{de2008z3} 
to check the satisfiability of the conjunction,
as done in~\cite{raman2013towards}.
If the constraint $\square \syssafetybool{i}$ is violated,
we provide the following \gls{nl} feedback:
``the skills in $\strategylabelout(\statevar)$ violate the safety constraint $\square \syssafetybool{i}$" (Line~\ref{line:safety_analysis_feedback}).
Fig.~\ref{fig:casestudy}F
presents an example safety analysis feedback for Example~\ref{example:1}.

\begin{algorithm}[t]\small
\caption{\textbf{Safety Analysis}}\label{alg:safety_analysis}
\KwIn{
Counterstrategy $\counterstrategy$,
System safety constraints $\syssafety = \bigwedge_i \square \syssafetybool{i}$
        }
\KwOut{\gls{nl} feedback on violated safety constraints}
\For{$\statevar, \statevar' \in \strategystates$ such that
$\statevar'\in \strategytranssys(\statevar, \strategytransenv(\statevar))$
and 
$\strategytranssys(\statevar', \strategytransenv(\statevar')) = \emptyset$}{\label{line:identify sink state}
\For{$\square \syssafetybool{i} $ in $\syssafety$}{\label{line:safety}
\If{
$\strategylabelinp(\statevar) \cup
\strategylabelout(\statevar) \cup
\bigcirc (\strategylabelinp(\statevar') \cup
\strategylabelout(\statevar')) \not\models
\syssafetybool{i}$
}{\label{line:sat_check}
\textbf{yield to feedback}: ``skills in $\strategylabelout(\statevar)$ violate the hard constraints $\square \syssafetybool{i}$."\label{line:safety_analysis_feedback}
}
}
}
\end{algorithm}


\textbf{Liveness analysis}.
Given the counterstrategy $\counterstrategy$ and 
the system liveness goals 
$\syslive = \bigwedge_{i=1}^n\square\lozenge \syslivebool{i}$,
we provide feedback on which liveness goals are not satisfied given the new skills. 
Algorithm~\ref{alg:live_analysis} describes our approach.
We first leverage Tarjan's algorithm~\cite{tarjan1972depth} to identify a set of strongly connected sets in 
the nodes $\strategystates$ of 
$\counterstrategy$ (Line~\ref{line:tarjan}).
We then only consider the sinking ones, i.e. there is no outgoing edge towards other components, and are not singletons without transitions (Line~\ref{line:sink_scc}).
Thus, we identify a sinking strongly connected component that can trap the robot within it once any state of the component is reached. 
We then identify the indices of a set of liveness goals $L$ satisfied in the component 
(Line~\ref{line:L}).
We note that 
the inequality comparisons are modulo $n$
(where $n\in \naturals$ is the total number of system liveness goals arranged in a cyclic order);
that is,
the condition
$\strategylabelgoal(\statevar) \le i < \strategylabelgoal(\statevar') \mod n$
in Line~\ref{line:L}
means that 
$\strategylabelgoal(\statevar) \le i < \strategylabelgoal(\statevar')$
if $\strategylabelgoal(\statevar) \le \strategylabelgoal(\statevar')$,
and $(\strategylabelgoal(\statevar) \le i \le n) \lor (1 \le i < \strategylabelgoal(\statevar'))$
if $\strategylabelgoal(\statevar) > \strategylabelgoal(\statevar')$.
%
Next, we obtain the indices of another set of liveness goals $L'$ that 
the robot attempts to, but cannot, satisfy in the component 
(Line~\ref{line:L_prime}).
Lastly, we provide feedback which states that the robot cannot satisfy the liveness goals in $L'$ after satisfying $L$
(Line~\ref{line:live_analysis_feedback}).
Fig.~\ref{fig:casestudy}G
presents an example feedback from liveness analysis for Example~\ref{example:1}.

Both 
the safety and liveness 
analyses have computation complexity 
$O(|\strategystates|^2)$
since they only access each state and transition of the counterstrategy at most a finite number of times.
Our feedback allows \gls{llms} to correct their mistakes and generate more promising repair candidates.


\begin{algorithm}[t]\small
\caption{\textbf{Liveness Analysis}}\label{alg:live_analysis}
\KwIn{
Counterstrategy $\counterstrategy$,
Liveness goals $\syslive = \bigwedge_{i=1}^n\square\lozenge \syslivebool{i}$
        }
\KwOut{\gls{nl} feedback on unsatisfied liveness goals}
$\tt{SCCs} \coloneqq \tt{find\_strongly\_connected\_components}(\counterstrategy)$\label{line:tarjan}\;
\For{$S\in \tt{SCCs}$}{\label{line:each_scc}
    \If{$(\forall \statevar \in S, \not\exists \statevar' \not\in S. \statevar'\in \strategytranssys(\statevar, \strategytransenv(\statevar))) \land (\exists \statevar \in S. \strategytranssys(\statevar, \strategytransenv(\statevar)) \neq \emptyset$)}{\label{line:sink_scc}
        $L \coloneqq \{i \mid \exists \statevar, \statevar' \in S. \statevar' \in \strategytranssys(\statevar, \strategytransenv(\statevar)),
        \strategylabelgoal(\statevar) \le i < \strategylabelgoal(\statevar') \mod n \}$\label{line:L}\;

        $L' \coloneqq \{i \mid \exists \statevar\in S. \strategylabelgoal(\statevar) = i\} \setminus L$ \label{line:L_prime}\;
        \Return ``The new skills cannot satisfy liveness goals 
        $\{\square\lozenge\syslivebool{i} \mid i\in L'\}$ 
        after satisfying liveness goals $\{\square\lozenge\syslivebool{i} \mid i\in L\}$."\label{line:live_analysis_feedback}
    
    }
}
\end{algorithm}

\section{Examples}\label{sec:experiments}

Our examples aim to explore the following questions:
\begin{enumerate}[label=\textbf{Q\arabic*}., ref=\textbf{Q\arabic*}]
    \item\label{question:baseline} 
            Does
            \framework 
            repair 
            violations
            more efficiently 
            than 
            the
            state-of-the-art
            formal-methods-based approach?

    \item\label{question:informalization} 
            How does each component (informalization, feedback) contribute to
            the overall performance of \frameworknospace?
    
\end{enumerate}

\begin{figure}[t] 
  \centering 
  \includegraphics[width=0.486\textwidth]{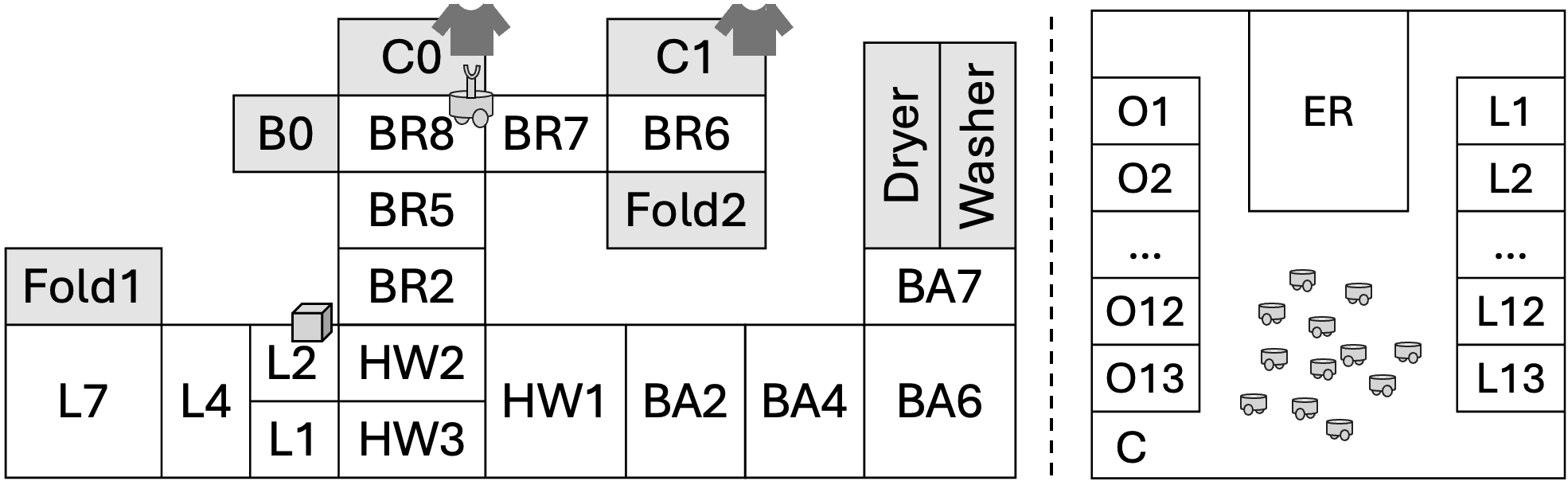} 
  \vspace{-0.25in}
  \caption{\textbf{Example workspace}.
  Left: An apartment-like workspace where a mobile manipulator performs laundry tasks.
  Right: An emergency room-like workspace where a robot swarm performs patrolling tasks.
  }
  \label{fig:example_workspace} 
  \vspace{-0.2in}
\end{figure}
\textbf{Examples}.
We evaluate 12 violations 
in four workspaces
over two abstraction types (object-centric and swarm~\cite{moarref2020automated}).
\begin{flushitemize}
    \item \textbf{Factory}. 
        A mobile manipulator operates in a factory-like workspace 
        (Fig.~\ref{fig:factory_cup}).
        During the execution, the robot encounters 
        three violations:
        \textit{factory-cone}
        where an obstacle \textit{cone} moves to the front of the robot,
        \textit{factory-cup} 
        where the status of a \textit{cup} changes
        (Example~\ref{example:1}),
        and \textit{factory-stone}
        where a static obstacle \textit{stone} unexpectedly moves.
        Detailed descriptions can be found in Sec. VC of ~\cite{meng2024automated}.
        
    \item \textbf{Large}. 
        A larger version of the factory example with additional regions surrounding the workspace in Fig.~\ref{fig:factory_cup}, increasing the number of input propositions from 29 to 71. 
        The three violations, \textit{large-cone}, \textit{large-cup}, and \textit{large-stone}, are the same as those in the factory example.
    \item \textbf{Laundry}. 
    A mobile manipulator is responsible for laundry tasks in an apartment-like workspace
    (Fig.~\ref{fig:example_workspace} left).
    In the violation \textit{laundry-box},
    an obstacle \textit{box} blocks the path to a folding table.
    In the violation \textit{laundry-pick}, 
    the skill that picks up a cloth fails, and the robot should attempt to pick up another cloth instead.
    In the violation \textit{laundry-fold}, 
    a folding table is malfunctioning, and the robot should use another folding table instead. 
    
    \item \label{example:er} \textbf{ER}. 
    A swarm of robots is in an \gls{er}-like workspace
    (Fig.~\ref{fig:example_workspace} right).
    In the violation \textit{\gls{er}-skill},
    a low-level controller failure causes some robots to enter an undesired region.
    In the violation \textit{\gls{er}-patient},
    a patient enters the  \gls{er} unexpectedly,
    requiring
    the swarm to visit the \gls{er} in response.
    In the violation \textit{\gls{er}-operate},
    a patient appears in an operation room unexpectedly,
    forcing the swarm to patrol other operation rooms instead.
    
\end{flushitemize}

\begin{figure}[t]
  \centering
  \begin{minipage}[t]{0.486\textwidth}
  \vspace{0pt}
    \begin{tcolorbox}[
      colback=blue!5!white,      
      colframe=blue!75,
      width=\textwidth,
      boxrule=0.5pt,
      sharp corners
    ]
    \scriptsize\textbf{A. Strategy informalization example for object-centric abstraction.} \\
    \textbf{Inputs}: 
["p\_base\_x0", "p\_base\_x1", "p\_base\_x2", "p\_base\_x3"]\\
\textbf{Outputs}:
["skill0", "skill1"]
\\
\textbf{Task specification and strategy}: \textless omitted for space \textgreater \\
%
%
\textbf{Behavior}:
Initially, the robot's base is in x0, and the liveness goal `p\_base\_x0` which requires the base to repeatedly reach x0 is satisfied. The robot then executes skill0 to move to x1 and then x3, satisfying the liveness goal `p\_base\_x3` which requires the base to repeatedly reach x3. Next, the robot executes skill1 to move back to x0 via x1, satisfying the liveness goal `p\_base\_x0` again. This behavior continues indefinitely.
\\
\textbf{Explanation}:
%
"Initially, the robot's base is in x0, and the liveness goal `p\_base\_x0` which requires the base to repeatedly reach x0 is satisfied" corresponds to node 0.
"The robot then executes skill0 to move to x1 and then x3, satisfying the liveness goal `p\_base\_x3` which requires the base to repeatedly reach x3" corresponds to the transitions from node 0 to node 1, node 2, node 3, and then node 4. 
"Next, the robot executes skill1 to move back to x0 via x1, satisfying the goal `p\_base\_x0` again" corresponds to the transitions from node 4 to node 5, node 6, node 7, and then back to node 0.


    \end{tcolorbox}
    \vspace{-1.5em}
    \begin{tcolorbox}[
      colback=red!5,       
      colframe=red!75,
      width=\textwidth,
      boxrule=0.5pt,
      sharp corners
    ]
    \scriptsize\textbf{B. Strategy informalization example for swarm abstraction.} \\
    \textbf{Inputs}: 
["p\_swarm\_x0", "p\_swarm\_x1", "p\_swarm\_x2", "p\_swarm\_x3"]
\\
\textbf{Outputs}:
["skill0", "skill1", "skill2", "skill3"]
\\
\textbf{Task specification and strategy}: \textless omitted for space \textgreater \\
%
%
\textbf{Behavior}:
Initially, the robot swarm is in x0 and the liveness goal `p\_swarm\_x0 \& !p\_swarm\_x1 \& !p\_swarm\_x2 \& !p\_swarm\_x3` which requires the swarm to repeatedly stay in x0 only is satisfied. 
The swarm then executes skill0 to reach x1 and x2, and then executes skill1 to reach x3, satisfying the goal `p\_swarm\_x3 \& !p\_swarm\_x0 \& !p\_swarm\_x1 \& !p\_swarm\_x2` which requires the swarm to repeatedly stay in x3 only. 
Next, the robot executes skill2 to move to x1 and x2, and finally executes skill3 to reach x0, satisfying the first goal again. This behavior continues indefinitely.
\\
\textbf{Explanation}:
"Initially, the robot swarm is in x0 and the liveness goal `p\_swarm\_x0 \& !p\_swarm\_x1 \& !p\_swarm\_x2 \& !p\_swarm\_x3` which requires the swarm to repeatedly stay in x0 only is satisfied" corresponds to node 0 and 1.
"The swarm then executes skill0 to reach x1 and x2" corresponds to the transitions from node 1 to node 2, node 2 to node 3 or node 4, and node 3 or node 4 to node 5 and then node 6.
"and then executes skill1 to reach x3, satisfying the goal `p\_swarm\_x3 \& !p\_swarm\_x0 \& !p\_swarm\_x1 \& !p\_swarm\_x2` which requires the swarm to repeatedly stay in x3 only" corresponds to the transitions from node 6 to node 7, and then node 8 or node 9, and then node 10.
"Next, the robot executes skill2 to move to x1 and x2" corresponds to the transitions from node 10 to node 11 or node 12, and then to node 13, node 14, and node 15.
"and finally executes skill3 to reach x0, satisfying the first goal again" corresponds to the transitions from node 15 to node 16 or node 17, and then to node 0. 
    \end{tcolorbox}

\vspace{-0.1in}
       \caption{\textbf{Strategy informalization examples}.
       Example translations from strategy to its behavior used to form the strategy informalization prompt (Fig.~\ref{fig:casestudy}C).
  (A) is
  for object-centric examples,
  and (B) for the swarm example.\
%
  }
  \label{fig:inform_example}
  \vspace{-0.2in}
    
  \end{minipage}
  \end{figure}
\textbf{Informalization examples}.
For strategy informalization
(Sec.~\ref{sec:informalization}),
we use an example in Fig.~\ref{fig:inform_example}A for object-centric abstraction
and the one in Fig.~\ref{fig:inform_example}B for swarm abstraction.


\textbf{Baseline and Ablations}. 
Our previous work~\cite{meng2024automated}
provides
a synthesis-based baseline
that modifies the pre and post-conditions of existing skills 
to suggest new skills 
that repair the violations.
We also consider two ablations of \frameworknospace:
\textit{no-inform} where we do not perform informalization and instead directly include the abstraction and strategy in the repair prompt;
and
\textit{no-feedback}, where we perform the iterative prompting without feedback from formal analyses.

\textbf{Setup}.
We run our experiments on
a Linux machine 
with a 2.3GHz 8-core AMD Ryzen\textsuperscript{TM}-7 CPU. 
We use OpenAI o1-preview-2024-09-12
as the LLM for \frameworknospace.
We measure runtime in seconds.
We set the runtime limit to 3000 seconds.
For \framework and its ablations,
we set a maximum limit for feedback iteration to be 5 
for monetary cost control of \gls{llms}. 
We run 5 trials with different random seeds for the baseline
and 3 trials for \framework and its ablations
for each violation.

\begin{figure*}[t]
  \centering 
  \includegraphics[width=\textwidth]{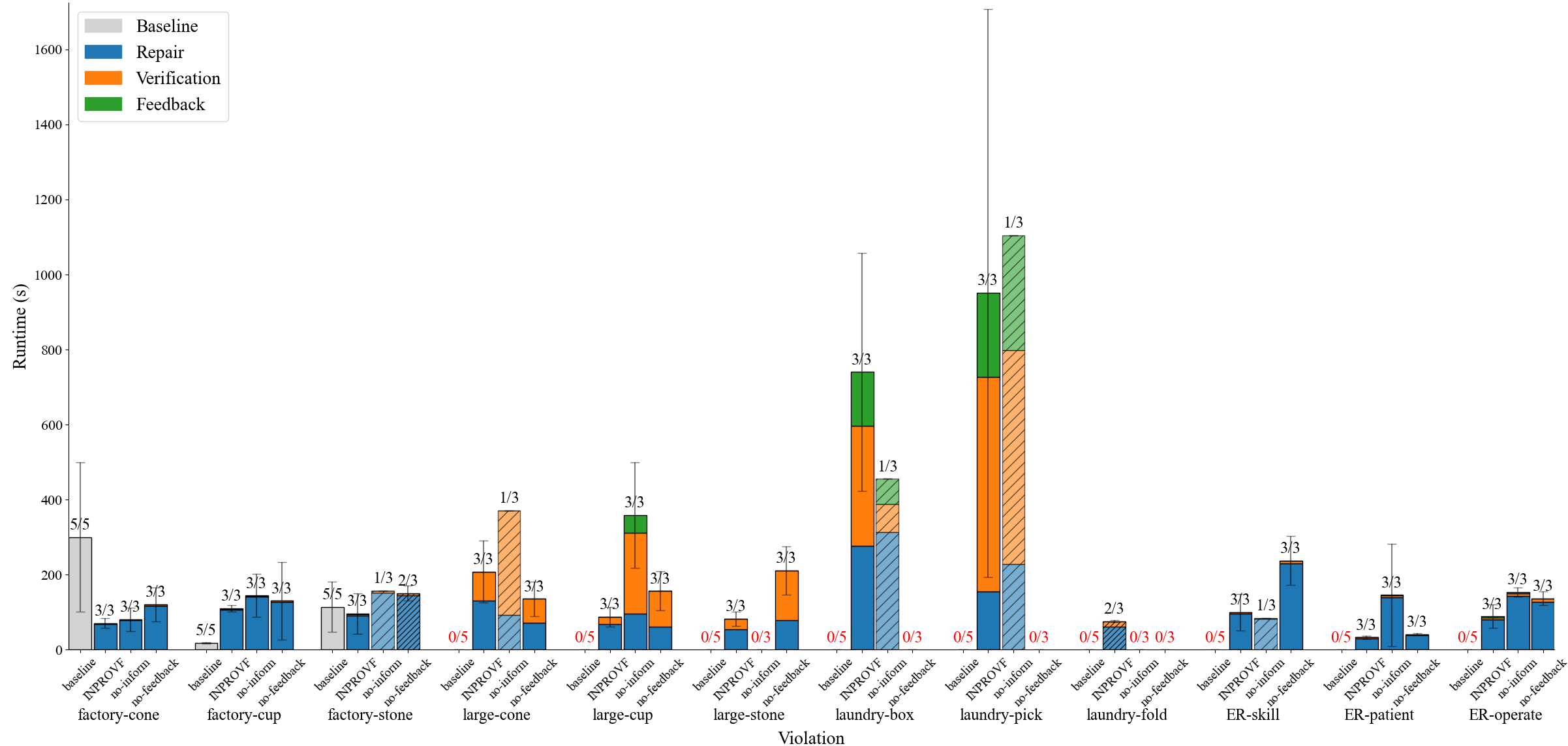}
  \caption{\textbf{Main results}. 
  Success rates and runtime comparison among 
  of baseline, 
  \frameworknospace,
  and ablation studies
  (no-inform and no-feedback).
  The success rates $n / m$ represents
  $n$ successful runs in $m$ total runs.
  We only show the runtime
  for the successful runs,
  including the detailed runtime breakdown of various components.
  Each error bar shows the standard deviation of the total runtime for each violation and configuration.
  %
  %
  }
  \vspace{-0.2in}
  \label{fig:main_results}
\end{figure*}

\textbf{Results}. 
Fig.~\ref{fig:main_results}
presents the success rates and the runtime comparison for the successful trails.
%
For \framework and its ablations, 
we also show the detailed runtime breakdown of each key component. 
We do not include the runtime for informalization
in Fig.~\ref{fig:main_results}
because we only perform informalization once offline
for each example;
other violations under the same example can reuse the informalization results.
Table~\ref{table:informalization_runtime}
shows the 
average runtime and standard deviation for each informalization component over three trials per example.

\textbf{Analysis}.
In the \textit{factory} example with a small state space (37 variables),
the average runtime of formal-methods-based baseline (144.05 s)
is comparable to that of \framework (92.36 s).
However, for the three larger examples (64-79 variables),
\framework significantly outperforms the baseline.
In these larger cases,
the baseline fails to complete within the 3000-second time limit.
We also remove the time limit for the \textit{larger-cup} violation,
but the baseline does not terminate after 
18523.93 s,
eventually causing a machine out-of-memory crash.
In contrast, 
for these larger examples, 
within five iterations, 
\framework successfully repairs every violation except one \textit{in a single run};
the failed run is for the \textit{laundry-fold} violation.
These results indicate that \framework is more scalable and efficient than the baseline in large state space (\ref{question:baseline}).
Our ablation studies show that \framework achieves a higher success rate (97.22\%) than \textit{no-inform} (55.56\%) and \textit{no-feedback} (72.22\%).
Additionally, in the successful runs,
the runtime for \textit{no-inform} is on average 7.43\% higher than that of \frameworknospace.
These results suggest both informalization and feedback are critical for
\framework to achieve high performance in runtime repair (\ref{question:informalization}).

\begin{table}[t]
\scriptsize
    \caption{Informalization average Runtime with standard deviation (s)
    }
    \centering
     \begin{tabular}{|c | c | c | c |}
         \hline
          & Abstraction-inform & Synthesis & Strategy-inform\\\hline
          Factory 
          & \roundto{2}{16.32294485304091} $\pm$ \roundto{2}{6.914429639619734} 
          & \roundto{2}{1.9401387373606365} $\pm$ \roundto{2}{0.5816800143654784}
          & \roundto{2}{65.79145153363545} $\pm$ \roundto{2}{29.108850645041777} \\\hline
          Large 
          & \roundto{2}{16.140743335088093} $\pm$ \roundto{2}{10.778117619768812} 
          & \roundto{2}{90.4338054921892} $\pm$ \roundto{2}{3.478558631999178} 
          & \roundto{2}{44.4648175769382} $\pm$ \roundto{2}{12.09099750665964}
          \\\hline
          Laundry 
          & \roundto{2}{17.827835241953533} $\pm$ \roundto{2}{7.777848659690155}
          & \roundto{2}{24.806466499964397} $\pm$ \roundto{2}{2.831337887265458} 
          & \roundto{2}{42.99164334932963} $\pm$ \roundto{2}{12.032999011902668}
          \\\hline
          ER 
          & \roundto{2}{10.172036727269491} $\pm$ \roundto{2}{4.642563774433939}
          & \roundto{2}{1.792496681213379} $\pm$ \roundto{2}{0.03345351596311696}
          & \roundto{2}{27.627652645111084} $\pm$ \roundto{2}{8.599099756774455}
          \\\hline
    \end{tabular}
    \label{table:informalization_runtime}
    \vspace{-0.15in}
\end{table}

\section{Limitations and Conclusion}
We present \frameworknospace, 
a novel hybrid framework that combines 
the scalability of \gls{llms}
and the guarantees of formal methods
to efficiently repair high-level robot controllers under assumption violations at runtime with prohibitively large state space.
Our experiments across various state space sizes 
indicate that our approach outperforms the formal methods-based approach in settings with large state spaces.

Nonetheless, 
our evaluation faces several limitations.
First,
since there is no standard benchmark for runtime repair,
our evaluation instances are limited.
In future work,
we plan to leverage \gls{llms} to procedurally generate a diverse range of repair instances across various abstraction types for a more comprehensive evaluation.
Moreover, 
the formal-methods-based repair executes on CPUs, and its implementation has not been optimized,
while the LLM components of \framework run on GPUs and benefit from extensive engineering optimizations.
Thus, a direct runtime comparison may not be fair.
Future work will focus on
developing 
evaluation metrics that better account for the hardware disparities.
Finally,
we currently
assess the quality of informalization indirectly
through the final repair correctness,
which is neither sound nor complete.
Future work will 
explore automatic evaluation methods that directly measure 
the quality of informalization.
Additionally,
we will work on
further
improving the scalability of formal methods components within our framework, 
such as realizability check.

\section*{Acknowledgments}
We thank Lin Luan for her invaluable assistance in enhancing the visual presentation of 
Figs.~\ref{fig:overview},~\ref{fig:framework},
and~\ref{fig:example_workspace},
Kevin Guan for his insightful discussion on the design of the experiments in Sec.~\ref{sec:experiments}, 
Bo-Ruei Huang and Andrew Violette for their insightful discussion on the ER example (Fig.~\ref{fig:example_workspace} right),
and anonymous reviewers from
ICLR 2025 VerifAI Workshop
for their detailed and insightful comments.

\bibliography{references}
\bibliographystyle{ieeetr}

\end{document}